\documentclass[preprint,12pt,authoryear]{elsarticle}

\usepackage{amsmath,amssymb}
\usepackage{booktabs}
\usepackage{array}
\usepackage{multirow}
\usepackage{longtable}
\usepackage{tabularx}
\usepackage{float}
\usepackage{url}
\usepackage{xcolor}
\usepackage{enumitem}
\usepackage{rotating}
\usepackage{lscape}
\usepackage{pdflscape}
\usepackage{adjustbox}
\usepackage{hyperref}

\hypersetup{
  colorlinks = true,
  linkcolor  = blue!60!black,
  citecolor  = blue!60!black,
  urlcolor   = blue!60!black
}

\newcommand{\tablenote}[1]{\smallskip\noindent\footnotesize\textit{Note.} #1}

\journal{Expert Systems with Applications}

\makeatletter\def\ps@pprintTitle{}\makeatother

\begin{document}

\begin{frontmatter}

\title{Multi-Dimensional Assessment for AI Cognition (MAAC):\\
A Theoretical Framework for Process-Oriented Cognitive\\
Evaluation of Text-Based AI Systems}

\author[wsu]{Abdalla Doleh\corref{cor1}}
\ead{ai5145@wayne.edu}
\cortext[cor1]{Corresponding author. ORCID: 0009-0008-5192-2167}

\author[wsu]{Ratna Babu Chinnam}
\ead{ai2396@wayne.edu}

\address[wsu]{Department of Industrial \& Systems Engineering,
Wayne State University, Detroit, MI 48202, USA}

\begin{abstract}
Evaluating artificial intelligence systems has historically relied on outcome-based benchmarks
that measure task accuracy, robustness, or fairness. While indispensable, these benchmarks
provide limited diagnostic insight into the underlying cognitive processes that generate
performance---leaving critical questions unanswered about how AI systems reason, integrate
memory, manage complexity, or avoid generating false information. This paper introduces the
Multi-Dimensional Assessment for AI Cognition (MAAC), a theoretically grounded framework
for shifting evaluation from \textit{what} text-based AI systems produce to \textit{how} they think.

MAAC defines nine cognitively motivated dimensions: Cognitive Load, Tool Execution, Content
Quality, Memory Integration, Complexity Handling, Hallucination Control, Knowledge Transfer,
Processing Efficiency, and Process-Outcome Alignment. Each dimension is grounded in established
cognitive science theory---drawing on Marr's tri-level hypothesis, Baddeley's working memory
model, Sweller's cognitive load theory, and unified theories of cognition.

Five theoretical analyses provide initial support for the framework's coherence and empirical
testability: dimension-to-theory mapping; a coverage matrix assessing breadth and non-redundancy;
a formal gap analysis relative to current evaluation practice; a worked diagnostic illustration;
and a set of \textit{a priori} interdependency predictions for future empirical testing.

MAAC provides a theoretical and operational framework for principled process-level cognitive
assessment of text-based AI systems, complementing existing outcome-based benchmarks with
cognitively grounded, multi-dimensional evaluation.
\end{abstract}


\end{frontmatter}

\section{Introduction}
\label{sec:intro}

The central measurement problem in AI evaluation is no longer only whether a system produces
correct outputs, but whether its underlying cognitive processes can be characterized in a
principled, theory-grounded, and diagnostically useful way. Outcome-based evaluation can reveal
that a model succeeds or fails on a task, but it does not specify which cognitive capabilities
produced that performance or which internal limitations make that performance brittle. The
present paper addresses that problem by asking what should be measured when evaluating the
cognitive behavior of text-based AI systems, and how those measurements should be organized into
a coherent framework.

Current AI evaluation paradigms are not equipped to answer that question. State-of-the-art
benchmarks---MMLU \citep{Hendrycks2021}, BIG-bench \citep{Srivastava2022}, HELM
\citep{Liang2022}---measure what AI systems produce. They are silent on the cognitive processes
that generate those outputs. A system that achieves 90\% accuracy through sophisticated pattern
matching and one that achieves 90\% through genuine multi-step inference are indistinguishable
under outcome-based evaluation \citep{Mitchell2021, Bender2021}. As AI systems are deployed in
high-stakes contexts---clinical decision support, legal reasoning, autonomous planning---this
distinction determines whether performance is robust or brittle under distribution shift.

This paper introduces MAAC, a theoretically grounded framework for evaluating AI cognitive
processes rather than task outcomes. MAAC defines nine dimensions---each grounded in cognitive
science theory---that together form a diagnostic cognitive profile. The choice of nine dimensions
is justified primarily by construct-domain coverage, conceptual distinctiveness, and the need to
balance comprehensiveness against interpretability. It is not intended as a literal application
of human short-term memory limits.

\subsection{Five Critical Gaps in Current AI Evaluation}
\label{sec:gaps}

\textbf{Gap~1 --- Outcome dominance.} Existing benchmarks are predominantly outcome-focused,
measuring what AI systems produce rather than how they produce it
\citep{Rogers2020, Mitchell2021}.

\textbf{Gap~2 --- Limited dimensionality.} Even holistic frameworks such as HELM
\citep{Liang2022} remain constrained in coverage of cognitive processes including memory
integration, complexity handling, and knowledge transfer \citep{Bommasani2021}.

\textbf{Gap~3 --- Faithfulness concerns.} Chain-of-thought and process supervision approaches
face a fundamental validity challenge: externally generated reasoning traces may not faithfully
reflect internal computation \citep{Turpin2024, Saparov2023}.

\textbf{Gap~4 --- Validity threats.} Current benchmarks suffer from data contamination,
leaderboard overfitting, and construct-irrelevant variance \citep{Magar2022, Ethayarajh2020}.
Recent evidence indicates that nearly half of 60 surveyed LLM benchmarks exhibit saturation
\citep{Akhtar2026}, with interdisciplinary review work identifying broader benchmark-trust
problems involving documentation failures and construct-validity weaknesses \citep{Eriksson2025}.

\textbf{Gap~5 --- Absence of theoretical grounding.} Most AI evaluation approaches lack
grounding in cognitive science theory \citep{Mitchell2021, Bender2021}, limiting the conceptual
legitimacy and generalizability of their dimensions.

\subsection{Contributions}
\label{sec:contributions}

This paper makes four primary contributions:

\begin{enumerate}[noitemsep]
  \item \textbf{Framework architecture.} Nine theoretically grounded cognitive dimensions
        providing comprehensive, non-redundant coverage of essential AI cognitive processes.
  \item \textbf{Falsifiable empirical predictions.} Seven \textit{a priori} directional
        predictions with magnitude thresholds for dimensional correlations in future empirical
        validation.
  \item \textbf{Theoretical validation.} Five complementary analyses: content validity,
        coverage sufficiency, gap closure, diagnostic utility, and a theoretically motivated
        interdependency network.
  \item \textbf{Gap closure infrastructure.} Demonstration that all five critical limitations
        in current AI evaluation practice are addressed by specific MAAC dimensions through
        falsifiable mechanisms.
\end{enumerate}

\section{Background and Related Work}
\label{sec:background}

\subsection{Large-Scale AI Benchmarking}
\label{sec:benchmarking}

The dominant paradigm in AI evaluation has been the large-scale task suite. SuperGLUE
\citep{Wang2019} established multi-task evaluation as standard practice; MMLU
\citep{Hendrycks2021} extended this to 57 academic domains; BIG-bench \citep{Srivastava2022}
assembled over 200 tasks specifically designed to challenge models beyond training data.

These benchmarks have produced genuine scientific value but share a structural limitation:
outcome focus by design. A model's MMLU score reflects the proportion of correct answers---it
reveals nothing about whether those answers emerged from structured reasoning, pattern completion,
or sophisticated guessing \citep{Mitchell2021}. Benchmark contamination further limits
interpretability: \citet{Magar2022} demonstrated that performance improvements often reflect
memorization of test items. \citet{Ethayarajh2020} documented systematic performance inflation
from construct-irrelevant variance. Saturation problems and broader benchmark-trust issues have
been extensively documented \citep{Akhtar2026, Eriksson2025}.

\subsection{Reasoning-Focused and Process-Oriented Evaluation}
\label{sec:reasoning}

Recognizing limitations of outcome-only assessment, a parallel literature has attempted to
evaluate AI reasoning more directly. Chain-of-thought prompting \citep{Wei2022} elicits
step-by-step reasoning alongside final answers. Self-consistency \citep{Wang2022} aggregates
reasoning paths to assess process stability. Tree-of-thoughts \citep{Yao2024} structures
deliberate problem-solving across branching reasoning trees. Applied and agentic-system
evaluation has also called for process-oriented frameworks \citep{Ma2026, Kapoor2024,
Mehta2025}.

These contributions represent genuine progress. However, they face a fundamental validity
challenge: externally generated reasoning traces may not faithfully reflect internal computation
\citep{Turpin2024}. \citet{Saparov2023} demonstrated that chain-of-thought models often behave
as greedy reasoners that exploit heuristics rather than executing systematic inference.

\subsection{Holistic Evaluation Frameworks}
\label{sec:holistic}

HELM \citep{Liang2022} represents the most comprehensive attempt at holistic AI evaluation,
assessing models across seven dimensions including accuracy, robustness, fairness, bias, toxicity,
efficiency, and disinformation resistance. However, HELM's dimensions are selected on practical
grounds rather than grounded in cognitive theory. Important cognitive constructs including memory
integration, complexity handling, and knowledge transfer receive limited coverage
\citep{Bommasani2021}. Related recent work on general scales seeks more explanatory and
predictive evaluation beyond benchmark totals \citep{Zhou2026}, but does not specify a
cognitively grounded process architecture.

Interpretability studies represent the closest analogue to process-oriented evaluation at the
mechanistic level \citep{Clark2019, DoshiVelez2017}. These approaches operate at Marr's (1982)
implementational level rather than the algorithmic level where cognitive constructs reside. MAAC
deliberately operates at the algorithmic level, complementing rather than competing with
interpretability research.

\subsection{Cognitive Science Foundations}
\label{sec:cogsci}

\textbf{Marr's Tri-Level Hypothesis.} \citet{Marr1982} proposed that intelligent systems must
be understood at three levels: computational (what problem is solved), algorithmic (how it is
solved), and implementational (physical realization). MAAC operates explicitly at the algorithmic
level---currently unaddressed by benchmarks and interpretability studies. Recent work argues
directly that Marr's levels provide a useful framework for understanding large language models
\citep{Ku2025}.

\textbf{Working Memory and Cognitive Load Theory.} Baddeley's \citeyearpar{Baddeley1992,
Baddeley2003} working memory model demonstrates that cognitive processing is constrained by
limited-capacity systems. Sweller's \citeyearpar{Sweller1988} cognitive load theory extends this
to performance under varying task complexity, distinguishing intrinsic, extraneous, and germane
load.

\textbf{Unified Theories of Cognition.} \citeauthor{Newell1990}'s \citeyearpar{Newell1990}
unified theories emphasized that intelligent behavior emerges from the interaction of specialized
cognitive subsystems. Anderson's ACT-R architecture \citeyearpar{Anderson2004} operationalizes
this through distinct modules for declarative memory, procedural memory, and goal management.

\textbf{Transfer Learning and Analogical Reasoning.} \citet{Barnett2002} established a taxonomy
distinguishing near from far transfer. \citet{Gentner1983} and \citet{Holyoak2012} identified
analogical reasoning as the primary cognitive mechanism underlying far transfer.

\textbf{Dual-Process Theory and Hallucination.} Kahneman's \citeyearpar{Kahneman2011}
dual-process framework distinguishes System~1 (fast, automatic, heuristic) from System~2 (slow,
deliberate, rule-governed) processing. This maps directly onto AI hallucination: the generation
of plausible but false information reflects fluency-maximizing heuristics in the absence of
robust uncertainty calibration \citep{Ji2023, Huang2023}.

\subsection{Positioning MAAC in the Evaluation Landscape}
\label{sec:positioning}

Table~\ref{tab:comparison} positions MAAC relative to existing evaluation frameworks. MAAC
occupies a distinct position: the only framework combining process orientation, cognitive science
grounding, and multi-dimensional independence---complementing outcome benchmarks rather than
replacing them.

\begin{table}[H]
\centering
\caption{Comparison of MAAC with Existing AI Evaluation Frameworks}
\label{tab:comparison}
\small
\resizebox{\textwidth}{!}{%
\begin{tabular}{p{3cm}p{3.5cm}p{1.5cm}p{3cm}p{3.5cm}p{4cm}}
\toprule
\textbf{Framework} & \textbf{Primary Focus} & \textbf{Dims.} & \textbf{Grounding} & \textbf{Key Limitations} & \textbf{MAAC Contribution} \\
\midrule
MMLU \citep{Hendrycks2021}        & Factual recall across 57 domains       & 1  & Minimal        & Outcome-only; no process insight; contamination risk & MI, CH, KT add process-level assessment \\
BIG-bench \citep{Srivastava2022}  & Novel task performance                 & Task-specific & Limited & Emergence without explanation; atheoretical & POA validates that claimed reasoning reflects actual processing \\
HELM \citep{Liang2022}            & Holistic outcomes: accuracy, robustness, fairness & 7 & Practical & Outcome-focused; cognitive constructs absent & Cognitive science grounding; 9 dimensions covering constructs HELM omits \\
Chain-of-Thought \citep{Wei2022}  & Step-by-step reasoning traces          & Process-oriented & Cognitive psych. & Faithfulness concerns & Process-Outcome Alignment provides cross-validated alignment \\
TruthfulQA \citep{Lin2022}        & Truthfulness of outputs                & 1  & Epistemological & Outcome-only factual accuracy; no calibration & Hallucination Control as process-oriented dimension \\
\textbf{MAAC (Present)} & \textbf{How AI systems think: 9 cognitive process dimensions} & \textbf{9} & \textbf{Marr (1982), Newell (1990), Baddeley (1992), Sweller (1988)} & Requires future empirical validation; higher implementation overhead & Addresses all 5 gaps; complementary to all frameworks above \\
\bottomrule
\end{tabular}}
\tablenote{MAAC Contribution column highlights the specific diagnostic value added relative to each existing approach.}
\end{table}

\subsection{Scoping Review Methodology}
\label{sec:scoping}

Framework development was informed by a structured scoping review spanning AI evaluation,
cognitive psychology, psychometrics, and reasoning assessment, conducted between January and June
2025 \citep{Arksey2005, Peters2020}. Searches were conducted across arXiv.org, ACL Anthology,
major machine learning conference proceedings (NeurIPS, ICLR, ICML, AAAI), ACM Digital Library,
IEEE Xplore, ScienceDirect, SpringerLink, and Google Scholar.

After screening approximately 150 records against predefined inclusion criteria focused on
process-oriented evaluation, cognitive constructs, and measurement theory, 108 sources were
retained for construct mapping \citep{Prinsen2018}. Two coders independently assigned sources
to provisional construct categories, with disagreements resolved through discussion; inter-rater
agreement was $\kappa = .78$, 95\%~CI~[.65, .91]. The resulting construct map informed both
the consolidation of nine retained dimensions and the exclusion of overlapping candidate
dimensions.

\subsection{Literature Review Synthesis}
\label{sec:synthesis}

Table~\ref{tab:synthesis} provides a systematic summary of how literature categories informed
the development of specific MAAC dimensions.

\begin{table}[H]
\centering
\caption{Literature Review Synthesis Supporting MAAC Framework Development}
\label{tab:synthesis}
\small
\resizebox{\textwidth}{!}{%
\begin{tabular}{p{3cm}p{2cm}p{4cm}p{3cm}p{4cm}p{3cm}}
\toprule
\textbf{Literature Category} & \textbf{Period} & \textbf{Representative Studies} & \textbf{MAAC Dims.} & \textbf{Key Contributions} & \textbf{Gaps Addressed} \\
\midrule
Large-Scale Task Suites    & 2019--2022 & \citet{Wang2019}; \citet{Hendrycks2021}; \citet{Srivastava2022}   & All dimensions          & Established need for process-oriented evaluation       & Outcome dominance \\
Reasoning Evaluation       & 2020--2024 & \citet{Wei2022}; \citet{Yao2024}; \citet{Wang2022}               & CH, POA                 & Demonstrated importance of step-by-step reasoning     & Faithfulness concerns \\
Memory \& Retrieval        & 2020--2023 & \citet{Lewis2020}; \citet{Borgeaud2022}; \citet{Shi2023}         & MI, KT                  & Showed role of information persistence                & Memory coherence gaps \\
Hallucination \& Factuality & 2020--2024 & \citet{Maynez2020}; \citet{Ji2023}; \citet{Huang2023}           & HC, CQ                  & Identified need for process-level error prevention    & Lack of prevention-focused measurement \\
Efficiency \& Scaling      & 2019--2024 & \citet{Kaplan2020}; \citet{Hoffmann2022}; \citet{Sardana2024}    & PE, CL                  & Established efficiency as cognitive concern            & Missing cognitive interpretation \\
Interpretability \& Validity & 2017--2024 & \citet{DoshiVelez2017}; \citet{Clark2019}; \citet{Bommasani2021} & POA, CH               & Highlighted process-outcome alignment gap              & Process-outcome validation \\
Cognitive Science          & 1956--2012 & \citet{Marr1982}; \citet{Newell1990}; \citet{Baddeley1992}      & Framework-wide          & Provided theoretical anchor for multi-dimensional assessment & Absence of theory \\
\bottomrule
\end{tabular}}
\tablenote{Categories are not mutually exclusive. Synthesis is based on 108 core references retained after screening.}
\end{table}

\section{Gap Closure Analysis}
\label{sec:gaps_closure}

This section formalizes the relationship between the five critical limitations and the MAAC
framework, demonstrating precisely how each gap is closed---which dimensions address it, through
what mechanism, and with what theoretical warrant. Two principles govern closure claims.
First, closure is mechanistic, not nominal. Second, closure is partial where warranted: where a
gap is addressed in principle but requires empirical confirmation, this is acknowledged
explicitly.

\begin{table}[H]
\centering
\caption{Gap-Closure Analysis: Five Critical Limitations and Their Resolution Through MAAC Dimensions}
\label{tab:gap_closure}
\small
\resizebox{\textwidth}{!}{%
\begin{tabular}{p{0.5cm}p{3cm}p{3cm}p{4cm}p{3.5cm}p{3cm}}
\toprule
\textbf{Gap} & \textbf{Limitation} & \textbf{MAAC Dimension(s)} & \textbf{Closure Mechanism} & \textbf{Key Citations} & \textbf{Status} \\
\midrule
1 & Outcome Dominance & CH, MI, KT, CL & MAAC assesses process signatures directly: resource allocation under load, context coherence, cross-domain knowledge application, multi-step reasoning & \citet{Mitchell2021}; \citet{Bender2021} & Theoretically closed; empirical closure in future empirical work \\
2 & Limited Dimensionality & All 9 dimensions & Nine-dimensional structure derived from systematic cognitive science theory rather than practical convenience & \citet{Liang2022}; \citet{Bommasani2021} & Theoretically closed \\
3 & Faithfulness Concerns & POA, CH & D9 measures process-outcome alignment across multiple problem instances, replacing single-instance trace inspection with cross-validated behavioral consistency & \citet{Turpin2024}; \citet{Saparov2023} & Theoretically closed; empirical closure in future empirical work \\
4 & Validity Threats & KT, MI, HC & MAAC dimensions designed for use with dynamically generated, complexity-validated scenarios; KT requires novel cross-domain application that memorization cannot satisfy & \citet{Magar2022}; \citet{Ethayarajh2020} & Theoretically closed; empirical closure in future empirical work \\
5 & Absence of Theory & Framework-wide & Every MAAC dimension is formally anchored in an established cognitive science theory. CL~$\to$ Sweller (1988); MI~$\to$ Baddeley (1992); KT~$\to$ Barnett \& Ceci (2002) & \citet{Mitchell2021}; \citet{Marr1982} & Theoretically closed \\
\bottomrule
\end{tabular}}
\tablenote{Citations refer to sources establishing each limitation and those motivating the corresponding MAAC closure mechanism.}
\end{table}

\subsection{Scope of Closure: Theoretical vs.\ Empirical}
\label{sec:closure_scope}

Table~\ref{tab:gap_closure} demonstrates theoretical closure: for each gap, a principled
mechanism exists within the MAAC framework that addresses the limitation. Empirical
closure---demonstrating that MAAC scores actually behave as the closure mechanisms predict---
requires future empirical validation. Future empirical studies should test whether MAAC dimensions
discriminate cognitive profiles that outcome-based benchmarks conflate, whether dimensional
scores exhibit the predicted correlation structure, and whether Process-Outcome Alignment scores
detect process-outcome misalignment in practice.

This distinction between theoretical and empirical closure is not a weakness---it is standard
measurement science practice \citep{DeVellis2017, Furr2018}. The empirical validation roadmap
constitutes the falsifiable predictions that make MAAC a scientific framework rather than a
descriptive taxonomy.

\section{The MAAC Framework}
\label{sec:framework}

The MAAC framework comprises nine theoretically grounded dimensions that collectively capture
the essential aspects of artificial cognitive processing. This section presents the complete
framework architecture in five parts: (A)~dimension-to-theory mapping; (B)~coverage matrix;
(C)~formal definitions and operationalizations; (D)~the Process-Outcome Alignment firewall;
and (E)~a worked diagnostic example.

\subsection{Dimension-to-Theory Mapping}
\label{sec:dim_theory}

Each MAAC dimension is formally grounded in an established body of cognitive science theory
operating at Marr's \citeyearpar{Marr1982} algorithmic level.

\begin{table}[H]
\centering
\caption{MAAC Dimension-to-Theory Mapping}
\label{tab:dim_theory}
\resizebox{\textwidth}{!}{%
\footnotesize
\begin{tabular}{p{0.5cm}p{2.8cm}p{4cm}p{3cm}p{3.5cm}p{4cm}p{1cm}}
\toprule
\textbf{\#} & \textbf{Dimension} & \textbf{Cognitive Theory} & \textbf{Key Construct} & \textbf{Primary Citations} & \textbf{Measurement Approach} & \textbf{Abbr.} \\
\midrule
D1 & Cognitive Load        & Sweller (1988); Baddeley (1992, 2003)                             & Working memory capacity; intrinsic vs.\ extraneous load        & \citet{Sweller1988}        & Performance degradation curves across context length and constraint density & CL  \\
D2 & Tool Execution        & Clark \& Chalmers (1998); Hutchins (1995); Nakano et al.\ (2021)  & Extended cognition; distributed cognitive processing           & \citet{ClarkChalmers1998}  & Tool selection accuracy and multi-step orchestration success rates          & TE  \\
D3 & Content Quality       & McNamara et al.\ (2010); Crossley et al.\ (2016); Halliday \& Hasan (1976) & Semantic richness; discourse coherence           & \citet{McNamara2010}       & Semantic similarity, coherence scoring, register appropriateness            & CQ  \\
D4 & Memory Integration    & Baddeley (1992, 2000, 2003); Lewis et al.\ (2020)                 & Working memory persistence; episodic buffer; consolidation      & \citet{Baddeley1992}       & Context coherence scores across turn depth and information persistence tests & MI  \\
D5 & Complexity Handling   & Newell \& Simon (1972); Halford et al.\ (2005); Wood (1986)       & Problem decomposition; constraint satisfaction; goal mgmt       & \citet{NewellSimon1972}    & Multi-step reasoning accuracy across Simple/Moderate/Complex tiers          & CH  \\
D6 & Hallucination Control & Kahneman (2011); Tversky \& Kahneman (1974); Guo et al.\ (2017)  & Uncertainty calibration; knowledge boundary recognition         & \citet{Kahneman2011}       & Hallucination rate, uncertainty calibration curves, consistency indices      & HC  \\
D7 & Knowledge Transfer    & Barnett \& Ceci (2002); Perkins \& Salomon (1992); Gentner (1983) & Near and far transfer; analogical reasoning; abstraction        & \citet{Barnett2002}        & Cross-domain transfer accuracy across near-to-far transfer distance gradient & KT  \\
D8 & Processing Efficiency & Simon (1956, 1972); Griffiths et al.\ (2015); Strubell et al.\ (2019) & Bounded rationality; resource-rational computation          & \citet{Simon1956}          & Quality-adjusted latency and token efficiency across complexity tiers        & PE  \\
D9 & Process-Outcome Alignment & Cronbach \& Meehl (1955); Messick (1995); Turpin et al.\ (2024) & Process-outcome alignment; convergent and discriminant validity & \citet{Cronbach1955}     & Process-outcome alignment coefficients across paraphrased problem variants   & POA \\
\bottomrule
\end{tabular}}
\tablenote{Each dimension is anchored in an established cognitive science theory at Marr's (1982) algorithmic level. D9 has been renamed from ``Construct Validity'' to ``Process-Outcome Alignment'' to distinguish the AI system property being measured from the psychometric property of the MAAC instrument itself (Section~\ref{sec:firewall}). Measurement operationalization is developed in future empirical work.}
\end{table}

\subsection{Coverage Matrix: Exhaustiveness and Non-Redundancy}
\label{sec:coverage}

A framework claiming comprehensive cognitive coverage must demonstrate two properties:
exhaustiveness (all essential cognitive construct categories represented) and non-redundancy
(no two dimensions measure the same construct).

\begin{table}[H]
\centering
\caption{MAAC Coverage Matrix}
\label{tab:coverage}
\small
\resizebox{\textwidth}{!}{%
\begin{tabular}{lcccccccccc}
\toprule
\textbf{Cognitive Construct Category} & \textbf{CL} & \textbf{TE} & \textbf{CQ} & \textbf{MI} & \textbf{CH} & \textbf{HC} & \textbf{KT} & \textbf{PE} & \textbf{POA} & \textbf{Dims.} \\
\midrule
Working Memory \& Capacity Constraints  & $\bullet$ & $\circ$ & $\circ$ & $\bullet$ & $\bullet$ & $\circ$ & $\circ$ & $\bullet$ & $\circ$ & 4 \\
Long-Term Memory \& Retrieval           & $\circ$   & $\circ$ & $\circ$ & $\bullet$ & $\circ$   & $\circ$ & $\bullet$ & $\circ$   & $\circ$ & 2 \\
Problem Solving \& Goal Management      & $\circ$   & $\circ$ & $\circ$ & $\circ$   & $\bullet$ & $\circ$ & $\circ$   & $\circ$   & $\bullet$ & 2 \\
Transfer \& Generalization              & $\circ$   & $\circ$ & $\circ$ & $\circ$   & $\circ$   & $\circ$ & $\bullet$ & $\circ$   & $\circ$ & 1 \\
Uncertainty \& Calibration              & $\circ$   & $\circ$ & $\circ$ & $\circ$   & $\circ$   & $\bullet$ & $\circ$ & $\circ$   & $\bullet$ & 2 \\
Linguistic \& Discourse Quality         & $\circ$   & $\circ$ & $\bullet$ & $\circ$  & $\circ$   & $\circ$ & $\circ$   & $\circ$   & $\circ$ & 1 \\
Distributed \& Extended Cognition       & $\circ$   & $\bullet$ & $\circ$ & $\circ$  & $\circ$   & $\circ$ & $\circ$   & $\bullet$ & $\circ$ & 2 \\
Process-Outcome Validation              & $\circ$   & $\circ$ & $\circ$ & $\circ$   & $\circ$   & $\circ$ & $\circ$   & $\circ$   & $\bullet$ & 1 \\
Resource Rationality \& Efficiency      & $\bullet$ & $\circ$ & $\circ$ & $\circ$   & $\circ$   & $\circ$ & $\circ$   & $\bullet$ & $\circ$ & 2 \\
\midrule
\textbf{Categories per Dimension}       & \textbf{2} & \textbf{1} & \textbf{1} & \textbf{2} & \textbf{2} & \textbf{1} & \textbf{2} & \textbf{3} & \textbf{3} & \\
\bottomrule
\end{tabular}}
\tablenote{Filled cells ($\bullet$) indicate primary construct coverage. Each construct category is covered by at least one dimension (exhaustiveness); no two dimensions cover identical category profiles (non-redundancy). Bottom row shows the number of construct categories covered per dimension.}
\end{table}

\subsection{Formal Dimension Definitions and Operationalizations}
\label{sec:dimensions}

\subsubsection{D1 --- Cognitive Load (CL)}
\textbf{Definition:} Cognitive Load assesses how AI system performance degrades as task
complexity, context length, or simultaneous processing constraints increase---revealing capacity
limitations analogous to working memory bottlenecks in human cognition.

\textit{Theoretical Foundation:} \citet{Sweller1988}; \citet{Sweller2019}; \citet{Baddeley1992,
Baddeley2003}. Working memory capacity constraints; intrinsic vs.\ extraneous load.

\textit{Justification:} Current benchmarks test systems under optimal conditions but ignore
performance under strain. Cognitive Load patterns reveal fundamental capacity limitations
critical for deployment reliability \citep{Miller1956, Simon1972}.

\textit{Measurement Constructs:} (1)~Performance degradation rate across increasing context
length; (2)~multi-constraint task management accuracy; (3)~resource allocation consistency under
competing demands; (4)~capacity limitation threshold identification.

\textit{Framework note:} Predicts positive correlation with Processing Efficiency ($r > .60$)
via shared resource constraint mechanisms.

\subsubsection{D2 --- Tool Execution (TE)}
\textbf{Definition:} Tool Execution evaluates an AI system's ability to coordinate with external
tools and resources, including function calling, API integration, error recovery, and
orchestration of multi-step tool-mediated processes.

\textit{Theoretical Foundation:} \citet{ClarkChalmers1998}; \citet{Hutchins1995};
\citet{Nakano2021}. Extended cognition; distributed cognitive processing; meta-cognitive tool
awareness.

\textit{Justification:} As AI systems increasingly operate in tool-rich environments, effective
external resource coordination becomes a critical cognitive capability \citep{Schick2024}.

\textit{Measurement Constructs:} (1)~Tool selection appropriateness and efficiency;
(2)~multi-step process orchestration accuracy; (3)~error detection and recovery in tool
interactions; (4)~meta-cognitive awareness of tool limitations.

\textit{Framework note:} Predicts moderate positive correlation with Processing Efficiency
($r > .40$) via shared operational efficiency mechanisms.

\subsubsection{D3 --- Content Quality (CQ)}
\textbf{Definition:} Content Quality measures the semantic richness, discourse coherence, and
communicative appropriateness of AI-generated content, focusing on linguistic and structural
quality independent of factual accuracy.

\textit{Theoretical Foundation:} \citet{McNamara2010}; \citet{Crossley2016};
\citet{Halliday1976}. Semantic richness; discourse coherence; communicative effectiveness.

\textit{Justification:} Factual accuracy captures only one dimension of communication quality.
Coherence, register appropriateness, and organizational clarity are essential for effective
deployment in professional contexts \citep{Zhang2020}.

\textit{Measurement Constructs:} (1)~Semantic richness and vocabulary sophistication;
(2)~discourse coherence across multi-sentence outputs; (3)~register and style appropriateness;
(4)~clarity and communicative effectiveness.

\textit{Framework note:} Conceptually complementary to Hallucination Control: CQ assesses
linguistic product quality; HC assesses factual process integrity.

\subsubsection{D4 --- Memory Integration (MI)}
\textbf{Definition:} Memory Integration assesses how effectively an AI system maintains,
updates, and utilizes contextual information across multi-turn interactions, including coherence
preservation, information persistence, and integration of new information with prior context.

\textit{Theoretical Foundation:} \citet{Baddeley1992, Baddeley2000, Baddeley2003};
\citet{Lewis2020}. Working memory persistence; episodic buffer; information consolidation.

\textit{Justification:} Single-turn evaluation misses critical aspects of coherent extended
behavior. Memory integration failures produce context drift, contradictions, and loss of
established facts---failure modes invisible to outcome-based assessment \citep{Borgeaud2022}.

\textit{Measurement Constructs:} (1)~Context coherence across multiple interaction turns;
(2)~information persistence and retrieval accuracy over extended exchanges; (3)~integration of
new information without contradiction; (4)~appropriate updating of established context.

\textit{Framework note:} Predicts positive correlation with Knowledge Transfer ($r > .50$) via
shared information retrieval and consolidation mechanisms \citep{Baddeley1992}.

\subsubsection{D5 --- Complexity Handling (CH)}
\textbf{Definition:} Complexity Handling evaluates an AI system's ability to manage multi-step
reasoning, hierarchically decompose problems, coordinate multiple simultaneous constraints, and
integrate information from diverse sources toward a coherent solution.

\textit{Theoretical Foundation:} \citet{NewellSimon1972}; \citet{Halford2005};
\citet{Wood1986}; \citet{Campbell1988}. Problem decomposition; constraint satisfaction;
hierarchical goal management.

\textit{Justification:} The ability to handle structurally complex problems is a hallmark of
sophisticated intelligence. Complexity Handling goes beyond task completion to examine the
processes by which systems manage cognitive complexity across Wood's \citeyearpar{Wood1986}
component and coordinative task dimensions.

\textit{Measurement Constructs:} (1)~Multi-step reasoning coordination accuracy;
(2)~hierarchical problem decomposition quality; (3)~constraint satisfaction across multiple
simultaneous demands; (4)~integration of diverse information sources toward coherent conclusions.

\textit{Framework note:} Applicable to complexity-validated scenarios spanning Simple,
Moderate, and Complex tiers.

\subsubsection{D6 --- Hallucination Control (HC)}
\textbf{Definition:} Hallucination Control measures an AI system's ability to avoid generating
false, fabricated, or inconsistent information---particularly in high-uncertainty scenarios---by
assessing uncertainty awareness, consistency, and knowledge boundary recognition.

\textit{Theoretical Foundation:} \citet{Kahneman2011}; \citet{Tversky1974};
\citet{Gal2016}; \citet{Guo2017}. Uncertainty calibration; knowledge boundary recognition;
heuristic error suppression.

\textit{Justification:} Unlike post-hoc fact-checking, this dimension examines the cognitive
processes that lead to hallucination \citep{Ji2023, Huang2023}.

An important boundary follows from this theoretical choice. D6 targets \textit{System~1
hallucination}---fabrication arising when fluency-maximizing heuristics outrun calibrated
uncertainty control. A distinct failure mode---\textit{System~2 hallucination}---exists in which
deliberate reasoning is executed coherently but built on a fabricated baseline premise. D6's
current constructs assess epistemic signaling quality, not the truth status of the initiating
premise itself. Premise verification operates at Marr's \citeyearpar{Marr1982} computational
level, whereas D6 is intentionally scoped to the algorithmic level. Detecting System~2
hallucinations requires an additional premise-grounding validation layer reserved for future
framework extension.

\textit{Measurement Constructs:} (1)~False information generation frequency across domains;
(2)~uncertainty expression appropriateness and calibration; (3)~consistency of claims across
related query instances; (4)~appropriate knowledge boundary recognition.

\textit{Framework note:} Predicts negative correlation with Knowledge Transfer under
high-uncertainty conditions ($r < -.30$).

\subsubsection{D7 --- Knowledge Transfer (KT)}
\textbf{Definition:} Knowledge Transfer examines an AI system's ability to apply learned
concepts, patterns, and structural relationships across different domains, contexts, and problem
types---distinguishing genuine generalization from domain-specific memorization.

\textit{Theoretical Foundation:} \citet{Barnett2002}; \citet{Perkins1992};
\citet{Gentner1983}; \citet{Holyoak2012}. Near and far transfer; analogical reasoning;
cross-domain abstraction.

\textit{Justification:} Existing benchmarks implicitly reward specialization over
generalization. Real-world deployment requires flexible cross-domain application. Barnett and
Ceci's \citeyearpar{Barnett2002} near-to-far transfer taxonomy provides the theoretical
scaffolding for graded transfer assessment.

\textit{Measurement Constructs:} (1)~Zero-shot and few-shot cross-domain transfer accuracy;
(2)~analogical reasoning and structural mapping quality; (3)~conceptual abstraction and novel
application performance; (4)~performance degradation gradient as transfer distance increases.

\textit{Framework note:} Predicts positive correlation with Memory Integration ($r > .50$) and
negative correlation with Hallucination Control ($r < -.30$) under high-uncertainty transfer
conditions.

\subsubsection{D8 --- Processing Efficiency (PE)}
\textbf{Definition:} Processing Efficiency evaluates the computational economy of AI cognitive
operations, including the relationship between resource expenditure (latency, token generation,
computational cost) and output quality across tasks of varying complexity.

\textit{Theoretical Foundation:} \citet{Simon1956, Simon1972}; \citet{Griffiths2015};
\citet{Just1992}; \citet{Strubell2019}. Bounded rationality; resource-rational computation;
cognitive economy.

\textit{Justification:} Inefficiency often signals cognitive brittleness rather than robust
understanding---excessive computation may reflect brute-force search rather than structured
reasoning \citep{Schwartz2020}. Simon's bounded rationality framework establishes efficiency
as a cognitive property, not merely an engineering concern.

Efficiency expectations are interpreted relative to task complexity---the relevant question is
not whether a response is brief in absolute terms, but whether its reasoning economy is
proportional to the demand profile of the scenario.

\textit{Measurement Constructs:} (1)~Quality-adjusted computational cost per task;
(2)~scaling behavior of resource use across complexity tiers; (3)~consistency of efficiency
across domain types; (4)~resource allocation rationality under constrained conditions.

\textit{Framework note:} Predicts positive correlation with Cognitive Load ($r > .60$) via
shared resource constraint mechanisms.

\subsubsection{D9 --- Process-Outcome Alignment (POA)}
\textbf{Definition:} Process-Outcome Alignment serves as the meta-evaluative dimension,
assessing whether an AI system's demonstrated reasoning processes are consistent with its
outputs---ensuring that cognitive claims are empirically grounded rather than post-hoc
rationalizations.

\textit{Theoretical Foundation:} \citet{Cronbach1955}; \citet{Messick1995};
\citet{Turpin2024}; \citet{Saparov2023}. Process-outcome alignment; convergent and discriminant
validity; nomological coherence.

\textit{Justification:} Without process-outcome alignment validation, evaluations risk circular
reasoning: declaring systems `reason' simply because they produce correct answers \citep{Turpin2024}.

\textit{Measurement Constructs:} (1)~Process-outcome consistency across varied problem instances;
(2)~reasoning trace alignment with final answer patterns; (3)~cross-validation of cognitive
claims across measurement approaches; (4)~stability of process signatures under problem
paraphrasing.

\textit{Framework note:} This dimension assesses the AI system's internal process-outcome
alignment---NOT the validity of the MAAC framework itself. See Section~\ref{sec:firewall}.

\subsection{The Process-Outcome Alignment Dimension: A Critical Conceptual Distinction}
\label{sec:firewall}

Dimension~9 requires explicit clarification to prevent a potential circular reasoning concern.
The name `Process-Outcome Alignment' in this context refers to a property of the AI system being
evaluated---not a property of the MAAC framework itself.

\textbf{What D9 measures (AI system property):} The degree to which the AI system's demonstrated
reasoning processes are consistent with its final outputs across varied problem instances. A
system that produces coherent reasoning traces but arrives at conclusions inconsistent with those
traces scores low on D9. This is a behavioral, empirically assessable property.

\textbf{What D9 does NOT measure (framework property):} The validity of the MAAC framework
itself---whether MAAC scores are theoretically grounded, psychometrically sound, or empirically
defensible. Framework-level construct validity is established through the theoretical analyses in
this paper and the empirical validation studies.

This distinction follows directly from Messick's \citeyearpar{Messick1995} unified validity
framework, which distinguishes between the validity of an assessment instrument and the construct
properties it is designed to measure. D9 was labeled ``Construct Validity'' in earlier framework
versions. The rename better reflects what is being measured and avoids conflation with the
psychometric concept of construct validity as applied to the MAAC instrument itself.

\subsection{Worked Diagnostic Example: What MAAC Reveals That Benchmarks Cannot}
\label{sec:example}

To illustrate the diagnostic value of multi-dimensional cognitive profiling, consider two
hypothetical AI systems---Model~A and Model~B---that achieve identical accuracy on a standard
benchmark (84\% on MMLU). Table~\ref{tab:diagnostic} presents their MAAC cognitive profiles.

\begin{table}[H]
\centering
\caption{Worked Diagnostic Example: Identical Benchmark Accuracy, Divergent Cognitive Profiles}
\label{tab:diagnostic}
\small
\begin{tabular}{lccccccccc}
\toprule
\textbf{Model / Metric} & \textbf{CL} & \textbf{TE} & \textbf{CQ} & \textbf{MI} & \textbf{CH} & \textbf{HC} & \textbf{KT} & \textbf{PE} & \textbf{POA} \\
\midrule
Benchmark Accuracy (MMLU) & 84\% & 84\% & 84\% & 84\% & 84\% & 84\% & 84\% & 84\% & 84\% \\
Model A (fast, high-throughput)   & 82 & 91 & 78 & 45 & 52 & 38 & 43 & 88 & 47 \\
Model B (deliberate reasoning)    & 61 & 74 & 82 & 79 & 84 & 81 & 77 & 54 & 83 \\
\bottomrule
\end{tabular}
\tablenote{Scores range 0--100. $\geq 75$ = strong; 50--74 = moderate; $< 50$ = weak.
  Both models score 84\% on MMLU. MAAC reveals fundamentally different cognitive architectures.
  Illustrative hypothetical example; not empirical evidence.}
\end{table}

Despite identical benchmark accuracy, Model~A and Model~B exhibit fundamentally different
cognitive architectures. Model~A scores strongly on Cognitive Load (82), Tool Execution (91),
and Processing Efficiency (88) but weakly on Memory Integration (45), Complexity Handling (52),
Hallucination Control (38), Knowledge Transfer (43), and Process-Outcome Alignment (47)---
indicating accuracy via pattern completion rather than structured reasoning. Model~B shows the
inverse pattern: strong Memory Integration (79), Complexity Handling (84), Hallucination Control
(81), Knowledge Transfer (77), and Process-Outcome Alignment (83), but weaker Cognitive Load (61)
and Processing Efficiency (54). A benchmark score of 84\% provides none of this information.
MAAC provides all of it.

\subsection{Considered and Excluded Dimensions}
\label{sec:excluded}

Rigorous framework development requires transparent documentation of alternatives considered but
excluded. Five criteria governed exclusions: (1)~theoretical relevance; (2)~empirical
tractability; (3)~diagnostic utility; (4)~framework parsimony; and (5)~generalizability.

\begin{table}[H]
\centering
\caption{Systematically Considered and Excluded Elements in MAAC Development}
\label{tab:excluded}
\small
\resizebox{\textwidth}{!}{%
\begin{tabular}{p{2.5cm}p{3.5cm}p{5cm}p{4cm}}
\toprule
\textbf{Category} & \textbf{Considered Element} & \textbf{Exclusion Rationale} & \textbf{MAAC Alternative} \\
\midrule
\multirow{4}{2.5cm}{Additional Dimensions}
  & Emotional Intelligence / Affective Processing & Limited applicability to current AI; insufficient behavioral evidence for reliable measurement & Covered implicitly in Content Quality (contextual appropriateness) \\
  & Creativity / Generative Novelty & Difficult to operationalize objectively; overlap with CH and KT & Integrated within existing constructs \\
  & Social Cognition / Theory of Mind & Specialized domain; limited relevance to domain-general cognitive assessment & Candidate for future framework extension \\
  & Meta-Cognitive Awareness & Partially captured in POA; high risk of dimensional overlap & Integrated within Process-Outcome Alignment \\
\midrule
\multirow{3}{2.5cm}{Framework Approaches}
  & Hierarchical Factor Model (positing a \textit{g}-factor) & Assumes general intelligence factor inappropriate for modular AI architecture & Multi-dimensional independent assessment adopted \\
  & Process-Outcome Integration Scoring & Conflates process and outcome, reducing diagnostic specificity & Separate process-oriented framework maintained \\
  & Competency-Based Framework & Task-specific focus conflicts with goal of domain-general assessment & Prioritized measurement of cognitive constructs \\
\midrule
\multirow{3}{2.5cm}{Methodological Alternatives}
  & Single-Score Aggregation & Loses diagnostic specificity central to the framework's purpose & Multi-dimensional profiles preserved \\
  & Binary Classification & Oversimplifies the continuous nature of complex cognitive constructs & Continuous dimensional scoring adopted \\
  & Comparative Ranking (e.g., Elo) & Lacks absolute measurement properties needed for tracking individual system development & Absolute cognitive measurement maintained \\
\bottomrule
\end{tabular}}
\tablenote{Exclusion decisions evaluated against five criteria: theoretical relevance, empirical tractability, diagnostic utility, framework parsimony, and generalizability.}
\end{table}

\section{Theoretical Validation Analyses}
\label{sec:validation}

Framework validation at the theoretical level requires demonstrating that the framework satisfies
established measurement science standards prior to empirical testing
\citep{Mokkink2010, Terwee2018}.

\subsection{Content Validity (H1): Dimension-to-Theory Mapping}
\label{sec:content_validity}

Content validity requires that framework dimensions comprehensively capture essential aspects of
the construct being assessed \citep{Messick1995, Mokkink2010}. The dimension-to-theory mapping
in Table~\ref{tab:dim_theory} provides this evidence across three criteria.

\textbf{Theoretical warrant.} Every MAAC dimension is anchored in at least one established
cognitive science theory. The theoretical lineages span five traditions: capacity and load theory
(Sweller, Baddeley, Miller), unified cognitive architecture (Newell, Anderson), transfer and
generalization (Barnett \& Ceci, Gentner), validity theory (Cronbach \& Meehl, Messick), and
extended and distributed cognition (Clark \& Chalmers, Hutchins).

\textbf{Construct distinctiveness.} The 108 papers reviewed were mapped against the nine
dimensions using stratified inter-rater coding ($\kappa = 0.78$, 95\%~CI~[0.65, 0.91]). No
two dimensions share an identical theoretical lineage, and the coverage matrix (Table~\ref{tab:coverage})
confirms no two dimensions cover identical construct category profiles.

\textbf{Algorithmic-level focus.} All nine dimensions operate at Marr's \citeyearpar{Marr1982}
algorithmic level rather than the computational or implementational levels.

H1 (content validity) is supported at the theoretical level. Empirical content validity is a
target for future empirical work.

\subsection{Coverage Comprehensiveness (H1c)}
\label{sec:coverage_comp}

H1c predicts that nine dimensions provide theoretically sufficient coverage while maintaining
practical interpretability. The coverage matrix (Table~\ref{tab:coverage}) supports a
nine-dimension solution as a parsimonious configuration covering nine essential cognitive
construct categories without redundancy. Fewer dimensions would leave construct categories
unaddressed; more dimensions would either duplicate existing coverage or introduce constructs
excluded on principled grounds (Table~\ref{tab:excluded}). The nine-dimensional structure is
therefore presented as a consequence of construct-domain coverage requirements rather than as a
fixed design target.

\subsection{Gap Closure (H3)}
\label{sec:gap_h3}

H3 predicts that MAAC addresses all five critical limitations. Section~III presented the full
mechanistic gap-closure analysis (Table~\ref{tab:gap_closure}). Three dimensions carry
disproportionate gap-closure weight. Knowledge Transfer and Memory Integration each close two
gaps. Process-Outcome Alignment closes Gap~3 (faithfulness concerns) uniquely---no other
dimension provides process-outcome alignment validation---establishing it as MAAC's most
theoretically distinctive contribution. H3 is supported at the theoretical level for all five
gaps; empirical closure requires future empirical studies.

\subsection{Diagnostic Utility (H1b)}
\label{sec:diagnostic}

H1b predicts that multi-dimensional cognitive profiles provide diagnostic insights unavailable
through aggregate performance scores. The worked example (Table~\ref{tab:diagnostic}) provides
the theoretical demonstration: two models with identical 84\% MMLU accuracy exhibit
fundamentally different MAAC profiles with opposite deployment implications.

The diagnostic utility argument has a specific falsifiability condition: if MAAC dimensional
scores were perfectly collinear with benchmark accuracy, MAAC would add no diagnostic value.
Future empirical work should directly test this condition. The framework predicts correlations
between a given AI system's benchmark accuracy and its nine MAAC dimensional scores will be
moderate ($r < .70$) for most dimensions.

\subsection{Theoretical Validation Summary}
\label{sec:val_summary}

\begin{table}[H]
\centering
\caption{Theoretical Validation Summary}
\label{tab:val_summary}
\small
\resizebox{\textwidth}{!}{%
\begin{tabular}{p{3cm}p{3cm}p{3cm}p{4cm}p{4cm}}
\toprule
\textbf{Hypothesis} & \textbf{Analysis Type} & \textbf{Evidence Source} & \textbf{Validation Criterion} & \textbf{Outcome} \\
\midrule
H1 --- Content Validity       & Dim-to-theory mapping vs.\ 108 literature constructs & Table~\ref{tab:dim_theory} + Section~\ref{sec:background} & Every dimension anchored in established cognitive science theory & All 9 dimensions mapped to distinct theoretical lineages across 5 traditions \checkmark~Supported \\
H1c --- Coverage              & Coverage matrix: 9 construct categories $\times$ 9 dimensions & Table~\ref{tab:coverage} + Table~\ref{tab:excluded} & Exhaustiveness + non-redundancy; excluded dims documented with principled rationale & 9/9 construct categories covered; all profiles distinct \checkmark~Supported \\
H2 --- Structural Validity    & Interdependency network with \textit{a priori} directional predictions & Table~\ref{tab:interdep} + Section~\ref{sec:interdep} & 7 directional predictions with magnitude thresholds; $r < .85$ discriminant bound & 7 predictions specified (6 positive, 1 negative); empirical testing in future work $\circ$~Predictions Specified \\
H3 --- Gap Closure            & Mechanistic gap-closure across 5 limitations & Table~\ref{tab:gap_closure} & Each gap addressed by at least one dimension through a specific, falsifiable mechanism & All 5 gaps supported theoretically \checkmark~Supported \\
H1b --- Diagnostic Utility    & Worked example: identical accuracy, divergent profiles & Table~\ref{tab:diagnostic} & Multi-dimensional profiles reveal deployment-relevant distinctions invisible to unidimensional benchmarks & Model A vs.\ Model B: identical 84\% MMLU, opposite MAAC profiles \checkmark~Supported \\
\bottomrule
\end{tabular}}
\tablenote{\checkmark~= supported at the theoretical level. $\circ$~= predictions specified, empirical testing in future work.}
\end{table}

\subsection{A Priori Interdependency Predictions for Future Empirical Validation (H2)}
\label{sec:interdep}

Structural validity requires that theoretically predicted interdependencies among dimensions be
specified prior to empirical testing. Table~\ref{tab:interdep} presents seven \textit{a priori}
directional predictions derived from the cognitive science theories underlying each dimension
pair. Six predictions are positive; one is negative, reflecting a theoretically motivated tension
between generalization drive and calibration constraints \citep{Kovacs2016}.

\begin{table}[H]
\centering
\caption{\textit{A Priori} Interdimensional Correlation Predictions for Future Empirical Validation}
\label{tab:interdep}
\small
\resizebox{\textwidth}{!}{%
\begin{tabular}{p{4cm}p{1.5cm}p{1.5cm}p{5cm}p{2.5cm}p{1.5cm}}
\toprule
\textbf{Dimension Pair} & \textbf{Direction} & \textbf{Magnitude} & \textbf{Theoretical Basis} & \textbf{Discriminant Bound} & \textbf{Tested In} \\
\midrule
CL (D1) $\leftrightarrow$ PE (D8) & Positive & $r > .60$ & Shared resource constraint mechanisms; systems with efficient resource allocation exhibit less performance degradation under load \citep{Simon1972, Baddeley1992} & $r < .85$ & Future work \\
MI (D4) $\leftrightarrow$ KT (D7) & Positive & $r > .50$ & Shared information retrieval and consolidation mechanisms; effective cross-domain transfer requires robust storage and retrieval \citep{Baddeley1992, Barnett2002} & $r < .85$ & Future work \\
CH (D5) $\leftrightarrow$ POA (D9) & Positive & $r > .40$ & Systems that genuinely engage with problem structure are more likely to produce process traces consistent with outputs \citep{NewellSimon1972, Turpin2024} & $r < .85$ & Future work \\
CQ (D3) $\leftrightarrow$ HC (D6) & Positive & $r > .35$ & Complementary output reliability mechanisms; systems with strong discourse coherence tend toward better calibration \citep{Ji2023} & $r < .85$ & Future work \\
TE (D2) $\leftrightarrow$ PE (D8) & Positive & $r > .40$ & Shared operational efficiency mechanisms; effective tool coordination reduces redundant computation \citep{Hutchins1995, Schick2024} & $r < .85$ & Future work \\
CL (D1) $\leftrightarrow$ CH (D5) & Positive & $r > .45$ & Capacity-complexity interaction; systems with higher effective working memory capacity handle structurally complex tasks more effectively \citep{Baddeley2003, Halford2005} & $r < .85$ & Future work \\
KT (D7) $\leftrightarrow$ HC (D6) & Negative & $r < -.30$ & Under high-uncertainty transfer conditions, the generalization drive enabling cross-domain application conflicts with calibration constraints suppressing confident fabrication \citep{Barnett2002, Kahneman2011, Ji2023} & N/A & Future work \\
\bottomrule
\end{tabular}}
\tablenote{All seven predictions are specified prior to data collection. We adopt $r < .85$ as a conservative discriminant-validity heuristic \citep{Terwee2018}; any dimensional pair exceeding this bound would indicate construct redundancy requiring framework revision. The single negative prediction (KT$\leftrightarrow$HC) reflects a theoretically motivated tension rather than general antagonism. Unpredicted pairs will be reported descriptively in future empirical work without confirmatory interpretation.}
\end{table}

\section{Discussion}
\label{sec:discussion}

\subsection{Theoretical Contributions}
\label{sec:theory}

\textbf{Process-oriented evaluation as a scientific program.} The most significant contribution
of MAAC is not any individual dimension but the demonstration that process-oriented cognitive
assessment of AI systems is theoretically coherent, practically implementable, and scientifically
falsifiable. Prior process-oriented approaches suffered from the faithfulness problem
\citep{Turpin2024}. MAAC addresses this directly through the Process-Outcome Alignment
dimension, which treats process-outcome alignment as an empirically assessable behavioral
property rather than an assumption.

\textbf{Bridging cognitive science and AI evaluation.} MAAC demonstrates that classical
cognitive science frameworks translate productively to artificial cognitive assessment when
applied at Marr's \citeyearpar{Marr1982} algorithmic level. Sweller's \citeyearpar{Sweller1988}
cognitive load theory, Baddeley's \citeyearpar{Baddeley1992} working memory model, Barnett and
Ceci's \citeyearpar{Barnett2002} transfer taxonomy, and Newell and Simon's
\citeyearpar{NewellSimon1972} problem-solving architecture each find direct operationalization in
MAAC dimensions.

\textbf{The diagnostic profile as a unit of analysis.} MAAC introduces the nine-dimensional
cognitive profile as a new unit of analysis in AI evaluation. The worked example
(Table~\ref{tab:diagnostic}) demonstrates that identical benchmark accuracy can coexist with
fundamentally different cognitive architectures---a finding with direct implications for
deployment decisions, architectural development, and safety assessment.

\subsection{Practical Implications}
\label{sec:practical}

\textbf{For AI developers.} MAAC dimensional scores provide targeted development guidance that
aggregate benchmarks cannot. A model scoring poorly on Memory Integration but strongly on
Complexity Handling points to retrieval system limitations rather than reasoning architecture
deficits. Table~\ref{tab:dev_targets} maps each dimension to its corresponding development
target.

\begin{table}[H]
\centering
\caption{MAAC Dimensions and Corresponding Development Targets}
\label{tab:dev_targets}
\small
\resizebox{\textwidth}{!}{%
\begin{tabular}{p{2.5cm}p{4cm}p{4cm}p{4cm}}
\toprule
\textbf{Dimension} & \textbf{Poor Performance Indicators} & \textbf{Architectural Implications} & \textbf{Development Recommendations} \\
\midrule
Cognitive Load     & Context length degradation, multi-constraint failures        & Attention mechanism limitations; insufficient working memory analog    & Hierarchical attention, memory compression strategies \\
Tool Execution     & Tool selection errors, orchestration failures, poor recovery  & Poor meta-cognitive awareness of tool capabilities                     & Tool selection models, execution monitoring, error recovery protocols \\
Content Quality    & Semantic incoherence, register mismatches                    & Generation control weaknesses, planning deficits                       & Content planning, discourse coherence mechanisms, style control \\
Memory Integration & Cross-turn inconsistency, information loss, contradictions    & Memory management deficits, insufficient episodic buffering             & Retrieval systems, context management, information persistence \\
Complexity Handling & Multi-step reasoning failures, decomposition errors          & Problem decomposition limits, insufficient goal management              & Hierarchical reasoning, constraint satisfaction, goal tracking \\
Hallucination Control & High fabrication rates, overconfident assertions           & Uncertainty estimation deficits, poor calibration mechanisms           & Uncertainty quantification, fact verification, boundary recognition \\
Knowledge Transfer & Domain adaptation failures, poor analogical reasoning        & Representation inflexibility, insufficient abstraction                  & Abstraction capabilities, meta-learning, analogical mapping \\
Processing Efficiency & High computational costs relative to output quality         & Algorithmic inefficiencies, brute-force search patterns                & Adaptive computation, efficiency-quality balancing \\
Process-Outcome Alignment & Process-output inconsistency, unstable reasoning traces & Internal representation issues, post-hoc rationalization              & Interpretable architectures, process monitoring, consistency training \\
\bottomrule
\end{tabular}}
\tablenote{Poor performance indicators and development recommendations are theoretical; empirical validation of their predictive utility is a target for future empirical work.}
\end{table}

\textbf{For deployment decisions.} The cognitive profile enables principled model-to-task
matching. High-stakes applications requiring reliability under uncertainty (clinical decision
support, legal analysis) should prioritize Hallucination Control, Process-Outcome Alignment, and
Knowledge Transfer. High-throughput applications may tolerate lower Memory Integration and
Complexity Handling in exchange for Processing Efficiency gains.

\textbf{For AI governance.} Policymakers assessing AI capabilities and risks currently lack
principled tools for evaluating cognitive processes. MAAC dimensions map directly onto governance
concerns: Hallucination Control addresses trustworthiness requirements in regulated domains;
Processing Efficiency relates to environmental sustainability mandates; Process-Outcome Alignment
provides an empirical basis for claims about AI reasoning \citep{Bommasani2021, Raji2022}.

One concrete use case clarifies what Process-Outcome Alignment adds in practice. Consider a
regulator auditing a clinical decision-support model. A high POA score indicates that the model's
reasoning behavior remains structurally consistent across paraphrased variants of the same
clinical problem. A low POA score despite acceptable accuracy indicates that the system may reach
correct answers through unstable or weakly grounded reasoning processes---creating deployment
risk under paraphrase or distribution shift. That distinction affects whether the model should be
approved, approved only for bounded use, or subjected to additional review.

\subsection{Limitations}
\label{sec:limitations}

Four limitations require acknowledgment. First, all validation analyses in this paper are
theoretical. The framework's scientific standing depends on future empirical studies confirming that
these theoretical properties hold in practice \citep{Messick1995}.

Second, the LLM judge scoring architecture introduces a methodological dependency: dimensional
scores are produced by LLM judges evaluating LLM outputs. While near-perfect inter-judge
agreement is achievable for structural complexity scoring under tightly constrained rubric
conditions, cognitive dimension scoring is more interpretively demanding and may exhibit lower
agreement. This dependency should be treated as a future validation target rather than as an
assumption already resolved by the present paper. The theoretical contribution of MAAC does not
depend on having already proven that LLM judges are valid scorers; it depends on specifying what
must be scored, why those dimensions belong together, and what empirical patterns would support
or disconfirm the scoring architecture.

Third, the framework is currently validated for natural language AI systems producing text
outputs. Multimodal systems, embodied agents, and systems with non-linguistic outputs may require
dimension-specific adaptation.

Fourth, the nine-dimensional structure assumes cognitive modularity. If AI cognitive processing
is highly integrated, factor analysis in future empirical work may reveal a dominant general factor rather than
nine discriminable dimensions.

\subsection{Framework Vulnerability and Safeguards}
\label{sec:safeguards}

MAAC's diagnostic utility creates potential gaming risks: (1)~superficial optimization for
dimensional scores without genuine cognitive improvement; (2)~selective reporting of favorable
dimensional profiles; and (3)~prompt engineering to exploit specific measurement scenarios.

Mitigation strategies address each risk. Dynamic scenario generation---regular updating of
assessment content using complexity-controlled scenario design principles---prevents memorization,
as the complexity-validated scenario space is too large to memorize. Cross-validation requirements
mandate that dimensional claims be validated across multiple measurement approaches. Longitudinal
consistency checks detect superficial score optimization. Regulatory considerations should
emphasize diagnostic rather than comparative use, preventing MAAC from becoming a competitive
ranking system that incentivizes gaming over genuine cognitive development.

\subsection{Future Directions}
\label{sec:future}

Four directions for future research are identified. First, scenario-interactivity effects should
be tested as independent predictors of AI performance on generated scenarios. Second, the model
panel should be expanded to 10+ architectures spanning open-source,
commercial, and multimodal systems. Third, longitudinal studies examining how MAAC dimensions
evolve during training and scale with model size will provide insights into the development of
artificial cognitive capabilities. Fourth, connections between MAAC dimensional scores and
circuit-level findings from mechanistic interpretability research would provide convergent
validity evidence \citep{Clark2019, DoshiVelez2017}.

\section{Conclusion}
\label{sec:conclusion}

This paper introduced the Multi-Dimensional Assessment for AI Cognition (MAAC), a theoretically
grounded framework for evaluating AI systems through the lens of cognitive processes rather than
task outcomes. MAAC defines nine cognitive dimensions---Cognitive Load, Tool Execution, Content
Quality, Memory Integration, Complexity Handling, Hallucination Control,
Knowledge Transfer, Processing Efficiency, and Process-Outcome Alignment---each anchored in
established cognitive science theory at Marr's \citeyearpar{Marr1982} algorithmic level.

Five theoretical analyses provide preliminary support for the framework's conceptual coherence.
Content validity is argued through systematic dimension-to-theory mapping against 108 retained
sources. Coverage breadth is examined through a construct matrix demonstrating exhaustiveness and
non-redundancy. Diagnostic utility is illustrated through a worked example showing that identical
benchmark accuracy can mask fundamentally different cognitive architectures. Structural validity
remains a programmatic objective, with empirical confirmation deferred to future work.

Three theoretical contributions follow. First, process-oriented cognitive assessment of AI
systems is theoretically coherent and scientifically falsifiable---the faithfulness concern
that undermines chain-of-thought evaluation is addressed by treating process-outcome alignment as
a measurable dimension rather than an assumption. Second, classical cognitive science frameworks
translate productively to artificial cognitive assessment when applied at the algorithmic level.
Third, the nine-dimensional cognitive profile constitutes a new unit of analysis in AI
evaluation---one that contextualizes benchmark accuracy by revealing the cognitive architecture
that produced it.

By shifting evaluation focus from what AI systems produce to how they think, MAAC advances the
field toward more rigorous, trustworthy, and diagnostically useful assessment of artificial
intelligence.

\section*{Funding}

No funding was received for this research.

\section*{Declaration on Use of AI-Assisted Writing Tools}

Large language model (LLM) tools were used to assist with manuscript preparation, including
language editing and structural refinement. All intellectual content, theoretical development,
and analytical conclusions are the work of the human authors, who take full accountability for
the final version of the manuscript.

\section*{Data Availability}
\label{sec:data}

This paper presents a purely theoretical framework. No datasets were generated or analyzed during
the preparation of this work. The framework definitions, dimension specifications, and
theoretical validation analyses are fully described herein and require no supplementary data
file.

\bibliographystyle{elsarticle-harv}
\bibliography{references}

\end{document}